# AgentMD: Empowering Language Agents for Risk Prediction with Large-Scale Clinical Tool Learning

Qiao Jin, M.D.[1], Zhizheng Wang, Ph.D.[1], Yifan Yang, B.Sc.[1,2], Qingqing Zhu, Ph.D.[1]
Donald Wright, M.D., M.H.Sc.[3], Thomas Huang, B.S.[3], W John Wilbur, M.D., Ph.D.[1]
Zhe He, Ph.D.[1,4], Andrew Taylor, M.D. M.H.S[3], Qingyu Chen, Ph.D.[3,1], Zhiyong Lu, Ph.D.[1,*]

[1]National Institutes of Health; [2]University of Maryland; [3]Yale University; [4]Florida State University

*Correspondence at: zhiyong.lu@nih.gov

## Abstract

Clinical calculators play a vital role in healthcare by offering accurate evidence-based predictions for various purposes such as prognosis. Nevertheless, their widespread utilization is frequently hindered by usability challenges, poor dissemination, and restricted functionality. Augmenting large language models with extensive collections of clinical calculators presents an opportunity to overcome these obstacles and improve workflow efficiency, but the scalability of the manual curation process poses a significant challenge. In response, we introduce AgentMD, a novel language agent capable of curating and applying clinical calculators across various clinical contexts. Using the published literature, AgentMD has automatically curated a collection of 2,164 diverse clinical calculators with executable functions and structured documentation, collectively named RiskCalcs. Manual evaluations show that RiskCalcs tools achieve an accuracy of over 80% on three quality metrics. At inference time, AgentMD can automatically select and apply the relevant RiskCalcs tools given any patient description. On the newly established RiskQA benchmark, AgentMD significantly outperforms chain-of-thought prompting with GPT-4 (87.7% *vs.* 40.9% in accuracy). Additionally, we also applied AgentMD to real-world clinical notes for analyzing both population-level and risk-level patient characteristics. In summary, our study illustrates the utility of language agents augmented with clinical calculators for healthcare analytics and patient care.

## Introduction

Clinical calculators have emerged as indispensable tools within healthcare settings, providing clinicians with evidence-based risk assessments essential for accurate diagnosis and prognostic evaluation[1,2]. As an example, the HEART score, a widely recognized tool, assists in evaluating the short-term risk of major adverse cardiac events based on points tallied from History, Electrocardiogram, Age, Risk Factors, and Troponin values, and has been shown to safely reduce admissions, imaging, and the financial burden in diverse populations[3]. Despite the success of clinical calculators to enhance efficiency and decision-making in healthcare, their adoption and utility are constrained by several factors. Clinicians must first recognize when and how to apply these tools, necessitating extensive knowledge of their functions—an issue compounded by slow dissemination. Clinical calculators are also frequently viewed as stand-alone tools, rarely combined together or applied at the same time. In addition, poor integration with Electronic Health Records (EHRs) interrupts clinical workflows, and without automatic data extraction from EHRs, clinicians are forced to input data manually. This not only hampers efficiency but also raises the risk of data entry errors. Moreover, the subjective nature of interpreting calculator components contributes to variability, compromising their consistency and undermining their overall reliability.

Language agents offer a promising approach to bridge this gap between clinical needs and existing risk calculators. These language agents[4], also known as Artificial Intelligence (AI) agents, are autonomous systems enabled by Large Language Models (LLMs) such as ChatGPT[5]. Unlike a vanilla language model, language agents can use external tools[6], including search engines[7,8], code interpreter[9], and domain-specific utilities[10-13] to reduce the potential hallucinations[14] in text generation. Prior efforts[15,16] have shown that LLMs augmented with calculators can significantly improve their efficacy in performing medical calculations. Furthermore, sophisticated LLMs, like GPT-4[17], are capable of creating versatile, reusable tools[18,19] (e.g., Application Programming Interfaces in Python) for other downstream language agents to use. Nevertheless, existing research predominantly addresses software engineering challenges, with minimal exploration into developing biomedical tools via LLMs. Additionally, constructing a comprehensive clinical calculator system that utilizes calculators published in the biomedical literature poses significant challenges, including curating the large array of published medical calculators and the early development stage of algorithms for automatic selection and utilization.

This study introduces AgentMD, a novel medical language agent framework designed to address two primary objectives: (1) the automated curation of a comprehensive library of medical calculators, enabling scalability across thousands of tools, and (2) the precise selection and application of these calculators to individual patient scenarios. Accordingly, the architecture of AgentMD encompasses two roles: as a tool maker, it automatically screens PubMed articles to identify relevant risk calculators, followed by the creation and validation of these tools, culminating in the assembly of a repository of structured risk calculator tools (RiskCalcs). In its role as a tool user, AgentMD employs an LLM-agnostic framework capable of selecting, computing, and summarizing the results from suitable risk calculators based on patient notes. To the best of our knowledge, this is the first work that generates, evaluates, and applies large-scale clinical tools with LLMs.

Since there have not been any prior efforts in automating the selection and application of large-scale clinical calculators, we manually create a new benchmark called RiskQA to evaluate the tool selection and usage capabilities of such systems. Evaluation results on RiskQA show that AgentMD can accurately select and use suitable clinical calculators, outperforming the compared models by a significant margin.

As a demonstration of its practical utility, we apply AgentMD to real-life clinical notes of Intensive Care Unit (ICU) patients from MIMIC-III and show that AgentMD can unveil insights into cohort characteristics at both population and individual risk levels. In summary, AgentMD exemplifies the promise of integrating language agents with clinical decision tools to improve risk predictions at scale, ultimately enhancing patient care outcomes.

## Results

### Automated Generation of RiskCalcs from PubMed Abstracts

As shown in Fig. 1a, AgentMD uses three steps to automatically curate risk calculator tools from PubMed: Screening, Drafting, and Verification. This process is detailed in Supplementary Figure 1. During article screening, AgentMD first applies a Boolean query "patient AND (risk OR mortality) AND (score OR point OR rule OR calculator)" to search for articles between January 2000 and April 2023 in PubMed. This results in 339,952 articles. AgentMD uses GPT-3.5-Turbo to screen them, leading to 33,033 articles that potentially describe a new risk score or calculator. Then, AgentMD uses GPT-4 to write structured medical calculators from these articles, where each calculator includes key sections such as Title, Purpose, Eligibility, Topics, Computing logic (with functions written in Python), Interpretation of results, and Utility. One example of the AgentMD-generated calculator in RiskCalcs is shown in Fig. 2a. Since the drafted calculators might contain inaccuracies such as hallucinations, AgentMD further uses GPT-4 to verify the generated calculators. The detailed prompts used at each step are shown in the Supplementary Materials.

Each calculator is classified by GPT-4 into one or more of the ten human organ systems in Medical Subject Headings (MeSH): Musculoskeletal (A02), Digestive (A03), Respiratory (A04), Urogenital (A05), Endocrine (A06), Cardiovascular (A07), Nervous (A08), Stomatognathic (A14), Hemic and Immune (A15), and Integumentary (A17) Systems. Fig. 1b shows the total number, PubMed article citation and population distributions of calculators in different systems. Overall, the cardiovascular system has the most calculators (811), and the stomatognathic system has the least (36). Although different systems differ in the number of calculators available, the distributions of total article citations and cohort sizes remain similar.

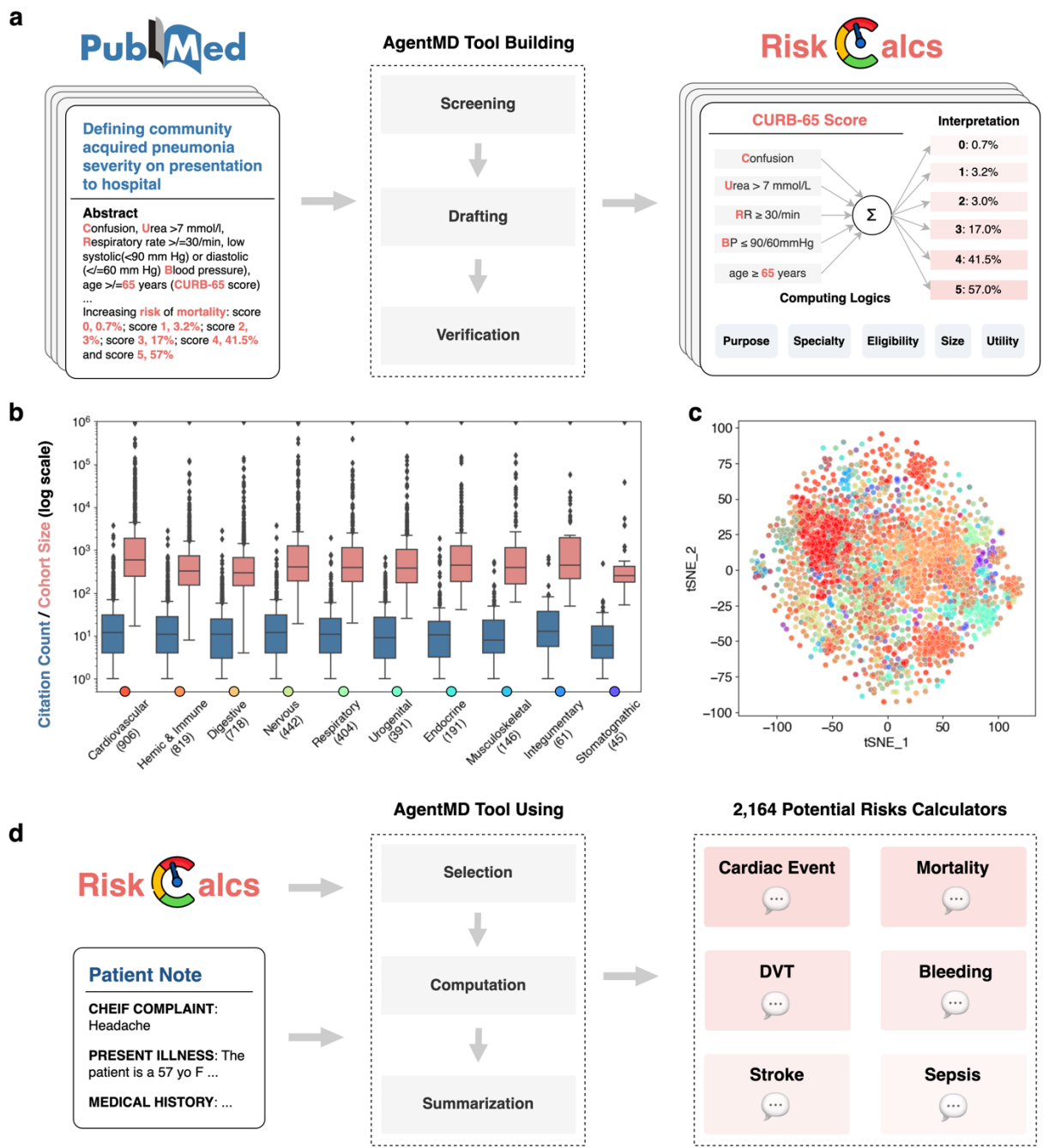

**Fig. 1. Overview of this study. a**, The process of building the RiskCalcs tool collection. AgentMD screens PubMed abstracts, drafts initial versions of the tools, and further verifies them. Only the verified calculators are included in RiskCalcs. **b**, The distribution of article citation count and calculator cohort size within the calculators of different organ

systems. **c**, t-SNE visualizations of the semantic representations of tools in RiskCalcs. **d**, AgentMD can apply the RiskCalcs tools to any given patient note. The process involves selection, computation, and summarization of eligible clinical calculators. AgentMD considers a wide variety of risks from 2,164 potential tools.

Fig. 1c shows the t-SNE visualizations[20] of the semantic representations of these calculators derived from MedCPT[21]. Each point represents a calculator and is color-coded by the organ system shown in Fig. 1b. The clustering results show that calculators of the same organ systems generally have similar semantic representations, suggesting the effectiveness of our topic classification system and the diversity of the clinical calculators in RiskCalcs.

**Applying RiskCalcs to Patient Notes by AgentMD**

Fig. 1d shows a high-level overview of how AgentMD computes risks from the 2,164 candidate calculators in RiskCalcs for a given patient note, including three main steps: selection, computation, and summarization. More technical details are presented in Fig. 2b. In the tool selection step, AgentMD first retrieves the top 10 most relevant calculators to the patient note using MedCPT, a foundation model for embedding biomedical texts, and then selects the eligible tool with LLMs. For each selected tool, AgentMD computes the patients' risk by generating Python code that calls the relevant and reusable tool functions in RiskCalcs. It is provided with a code interpreter to interact with – the Python interpreter returns the printed results or error messages when executing the code that

AgentMD writes to use the clinical tools. Based on the returned message from the code interpreter, AgentMD either re-tries other code actions or summarizes the whole interaction history into a paragraph of the risk calculation results. If the required parameters from the clinical calculator are missing from the patient notes, AgentMD will make a range estimation based on the best- and worst-case scenarios.

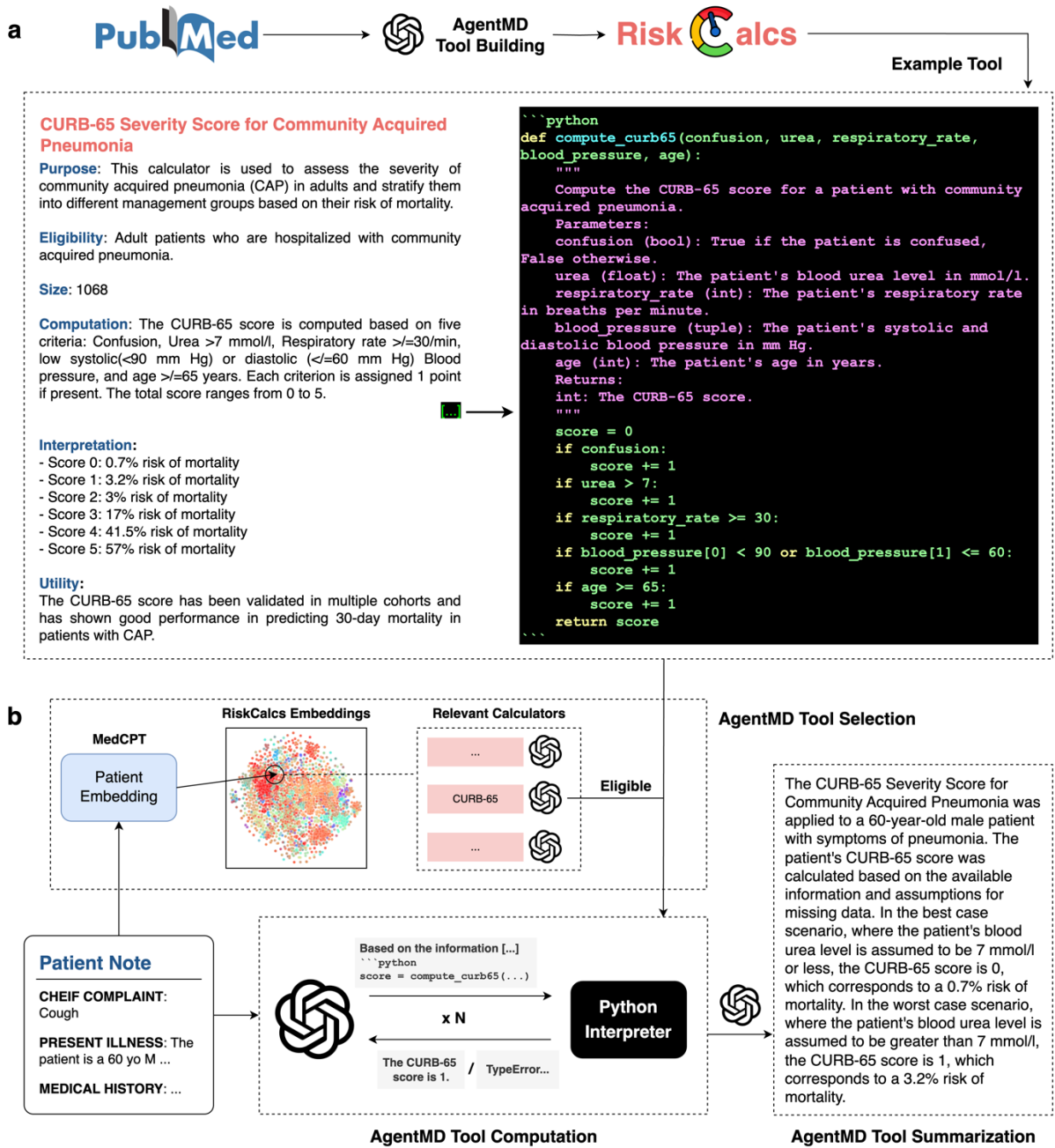

**Fig. 2. The methodology of AgentMD tool making and tool using**. **a**, An example of RiskCalcs calculator generated by AgentMD, based on the abstract of PMID: 12728155[22]. **b**, The methodology of AgentMD tool using, including tool selection, tool computation, and tool summarization. "" denotes the usage of LLMs such as GPT-4.

**RiskCalcs Tools Feature High Quality and Extensive Coverage**

Fig. 3 displays the evaluation results of the calculators in RiskCalcs. We manually evaluate two representative subsets of RiskCalcs: the top 50 most cited calculators (Fig. 2a); and a random sample of 50 other calculators (Fig. 2b). For each calculator, three annotators are employed to evaluate the quality, coverage, and question answering (QA) correctness of the tool. The consensus of three annotators is used as the ground-truth labels. The PubMed IDs of the calculators can be found in the Supplementary material.

The quality evaluation includes three aspects: raw abstract, computing logics, and result interpretations. Overall, the accuracies of the RiskCalcs tools with respect to the three quality aspects are high: The average correctness of the raw abstract, computing logics, and result interpretations are 87.0%, 87.6%, 89.0%, respectively. Specifically, 92.0% of the used abstracts contain relevant information (labeled as correct or partially correct) to write a risk calculator and only 8% of the abstracts are not suitable for writing a calculator. Among the 92.0% abstracts with relevant information, most (82.0%) contain sufficient details in computing logics and result interpretations, and only 10.0% abstracts miss certain information such as the interpretation of a score range. Similarly, most calculators have correct or partially correct computing logic (83% correct and 8% partially correct, 91.0% in total) and interpretation of results (83% correct and 12% partially correct, 95.0% in total). The most common cause of a partially correct computing logic and result interpretation is the lack of executable Python functions and missing certain result interpretations, respectively.

We also evaluated the coverage of the calculators by checking whether they have been implemented as an online tool. For this, we have searched the calculators in MDCalc[23] and Google. MDCalc is one of the largest online hubs containing over 700 clinical calculators. We also searched the first page returned from Google to find if the calculator has other online implementations. As expected, the majority (68.0%) of the top-25 most cited calculators in RiskCalcs have online implementations. However, the proportion is only 28.0% for calculators between the top 25-50. Risk calculators from many highly-cited studies, such as the Euro-EWING 99 trial[24], are not implemented by any websites but are automatically converted by AgentMD into a computable tool. Similarly, we did not find any online implementations for most (96.0%) randomly sampled calculators in RiskCalcs. Among the calculators with at least one online version, only 53.8% (14/26) have been implemented by both MDCalcs and other online sources, while the rest 46.2% have been only implemented in one source. This indicates that manual implementations of clinical calculators are limited in scale and lagging in progress. Overall, our coverage results show that RiskCalcs built by AgentMD can serve as a supplement resource of clinical calculators to the existing online hubs.

**Fig. 3. Quality and coverage analysis of RiskCalcs. a**, Evaluation results of the top-50 most cited calculators in RiskCalcs. **b**, Evaluation results of a random sample of 50 calculators in RiskCalcs. Abst.: abstract; Logic: computing logics; Interp.: result interpretation. Q1-Q5 denote five questions (clinical vignettes) generated for each tool.

In addition to the quality and coverage metrics, we also evaluated the accuracy of the computing logics of calculators in a well-controlled environment. For this, we used GPT-4 to generate five sets of potential parameter values (Q1-Q5 in Fig. 3) given the computing logics of each risk calculator. We provided AgentMD with both the calculator and the parameter set to compute the results, which we denote as AgentMD calculations. Then, we manually performed the result computation with the same set of parameters, either with the raw PubMed abstracts (internal validation) or with the online implementation (external validation) if available. Overall, only 8.4% (42/500) of the AgentMD calculations are inconsistent with manual calculations, while 17.4% and 74.2% of the AgentMD calculations are consistent with external and internal manual calculations, respectively. These results have further validated the accuracy of AgentMD computation and the quality of the RiskCalcs tools.

**AgentMD can accurately perform risk prediction tasks in RiskQA**

We evaluate AgentMD on RiskQA, a novel dataset introduced in this work following the format of multi-choice medical question answering that is commonly used to evaluate biomedical LLMs[25,26]. To construct the RiskQA dataset, we re-used the 350 manually validated sets of parameters for RiskCalcs calculators with correct computing logics and result interpretations. For each parameter set and validated calculation, we further used GPT-4 to expand them into a clinical vignette, possible choices, and the correct answer in the style of a United States Medical License Examination (USMLE) question (one example is shown in Fig 4a). Prior studies such as Articulate Medical Intelligence Explorer (AIME) also generated synthetic vignettes for evaluations[27]. The required calculator is not revealed in the RiskQA question so that the dataset can also test a system's capability in selecting the suitable tool. Unlike the computing logic evaluation shown above, RiskQA requires a system to select the suitable risk calculator, conduct correct computing, and provide appropriate interpretations.

Experimental results on RiskQA are shown in Fig. 4. When applied to this task, AgentMD first selects a tool from the RiskCalcs collection, then uses it to compute the risk for the given patient and predicts an answer choice, as shown in Fig. 4a. We first compare AgentMD with Chain-of-Thought (CoT)[28], a widely-used prompting strategy for LLMs. AgentMD surpasses CoT by 70.1% (0.546 vs. 0.321 in accuracy, Fig. 2b) and 114.4% (0.877 vs. 0.409, Fig. 2c) with GPT-3.5 and GPT-4 as the base model, respectively. Surprisingly, AgentMD with GPT-3.5 even outperforms CoT with GPT-4 (0.546 vs. 0.409). These results clearly demonstrate that large language models, when provided with a well-curated toolbox of clinical calculators, can accurately select the suitable calculator and effectively perform medical calculation tasks.

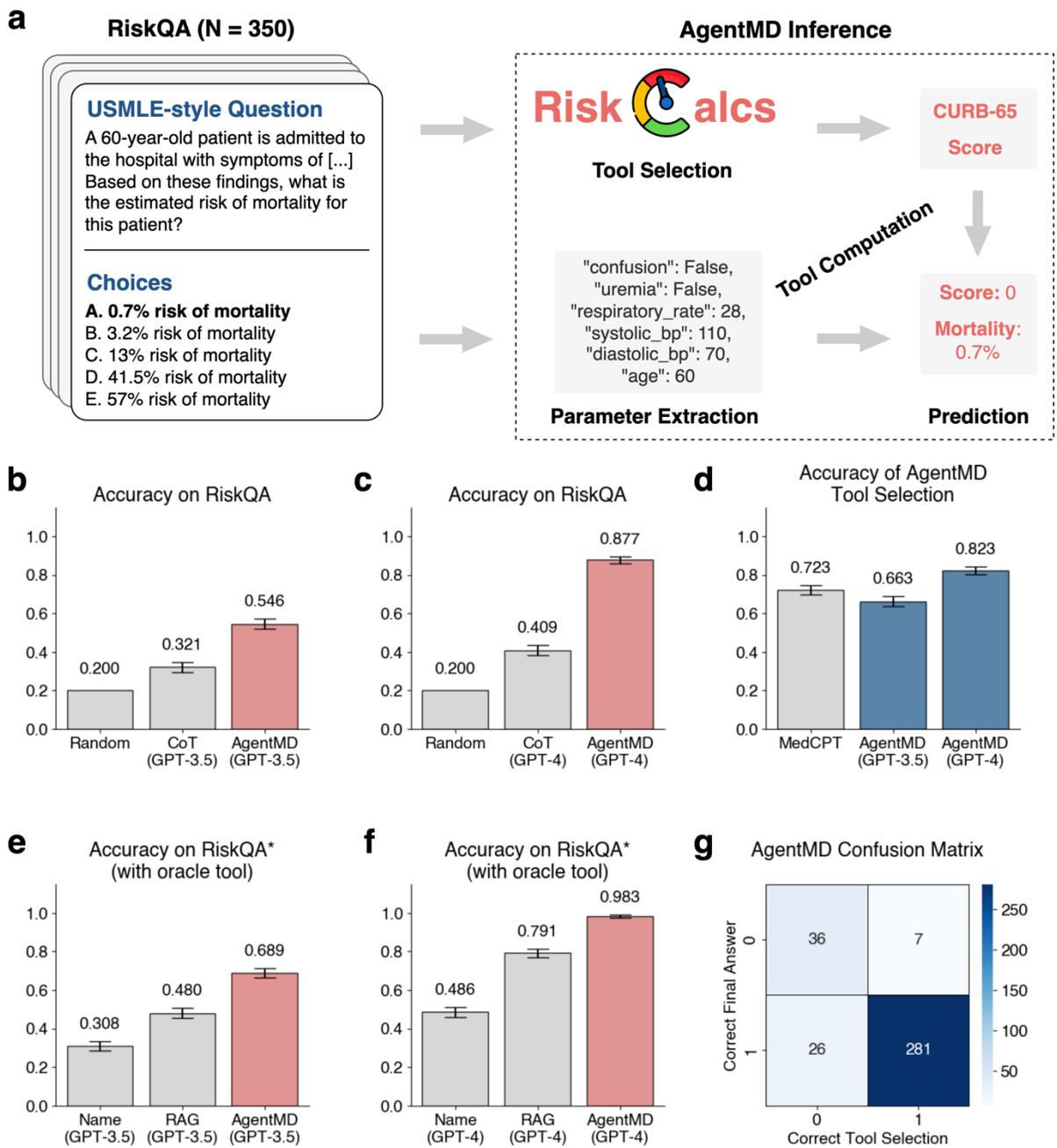

**Fig. 4. Evaluations of AgentMD on RiskQA. a**, Example of a question in RiskQA and how AgentMD answers it. **b**, The performance of GPT-3.5-based AgentMD compared to Chain-of-Thought (CoT) prompting on RiskQA. **c**, The performance of GPT-4-based AgentMD compared to CoT prompting on RiskQA. **d**, The accuracy of tool selection by

MedCPT and AgentMD. **e**, The performance of GPT-3.5-based AgentMD compared to Retrieval-augmented Generation (RAG) and Calculator-Name Prompting on RiskQA*, where the oracle calculator is provided. **f**, The performance of GPT-4-based AgentMD compared to RAG and Calculator-Name Prompting on RiskQA*. **g**, The confusion matrix of GPT-4-based AgentMD between the correctness of tool selection and the final answer.

Fig. 2d shows the accuracy of the intermediate tool selection step. As a baseline, dense retrieval with MedCPT achieves a top-1 accuracy of 0.723. On RiskQA, AgentMD selects the most suitable tool from the top 10 tools returned by MedCPT. Our results show that GPT-4-based AgentMD can better select the required tool than MedCPT, which is in turn better than GPT-3.5-based AgentMD. This highlights the importance of the backbone LLM when AgentMD conducts tool selection. The GPT-4-based AgentMD confusion matrix between the final choice correctness and the tool selection correctness is shown in Fig. 2g. Surprisingly, even with an incorrect tool selected, 41.9% of the final choices are correct (26/62), indicating a level of redundancy among the results of different clinical calculators.

We also evaluated different methods on RiskQA*, an easier version of RiskQA where the required risk calculator is provided in the question. Under this setting, AgentMD is directly provided with the oracle tool in the RiskCalcs. We compared AgentMD with a Retrieval-augmented Generation (RAG) method where the raw text abstract of the oracle calculator is provided. In addition, we also tested a setting where only the Name of the oracle calculator (e.g., "the CURB-65 score") is provided for the LLMs. The results on RiskQA* are shown in Fig. 4e and Fig. 4f, where GPT-3.5 and GPT-4 are used as the backbone LLMs, respectively. As expected, model performance under this RiskQA* is better than the original RiskQA. In both cases with GPT-3.5 and GPT-4, AgentMD is significantly better than RAG, which is in turn better than just providing the calculator title (the Name baseline). These results suggest that clinical calculators are better implemented with code snippets and utilized with a Python interpreter (as in AgentMD), in comparison to their raw textual descriptions as in previous studies like Almanac[16].

**AgentMD can be applied to clinical notes and provide cohort-level risk insights**

In this section, we apply AgentMD to 9,822 admission notes in MIMIC-III[29], a database of critical care electronic medical records. As shown in Fig. 5a, AgentMD first generates a list of potential risks with their quantitative likelihoods for each patient, where the detailed methodology is shown in Fig. 2. Then, we aggregate the AgentMD results by the 1,039 risk calculators that have been applied to the patients (Fig. 5b). For each calculator, AgentMD ranks the eligible patients by a set of metrics concerning the specificity, severity, urgency, and absence from the note. Fig. 5c shows the number of applied calculators per patient, which approximately follows a normal distribution with a mean value of 4.6. On the other hand, the number of eligible patients per tool follows a long-tail distribution, as shown in Fig. 5d.

Fig. 5e illustrates the distribution of patient results for the two most commonly applied calculators by AgentMD. The first calculator predicts the short-term mortality of acute exacerbation of chronic heart failure[30]. While the mean specificity is low, which indicates that most of the needed parameters are missing from the patient notes, its urgency and severity distributions have higher mean values. The absence distribution of the calculator has two peaks – the higher one close to 100 and the lower one close to 0 – which indicates that the short-term mortality is not assessed most of the eligible patient notes. The second calculator predicts 4-year mortality in older adults[31]. Unlike the short-term mortality, most patient results for the 4-year morality prediction are not urgent, and the severity also distributes differently with a lower mean value. As expected, they are mostly absent from the patient notes. These two examples demonstrate how different calculator results can provide distinct insights regarding the specific risks of the eligible population.

Fig. 5f shows the number of patients that have been applied to by calculators of various risk types. Overall, mortality is the most commonly considered risk among the ICU patients in the MIMIC-III dataset, with over 60% of patients (6,060 in 9,822) eligible for at least one mortality-related risk calculator. Cardiac events (3,174) and stroke risks (3,005) are also often calculated, which are followed by respiratory system events (2,284), bleeding (2,260), and infections (2,227). For each specific risk, one can visualize the calculation result distribution similar to Fig. 5e to study the individual risk-level cohort characteristics.

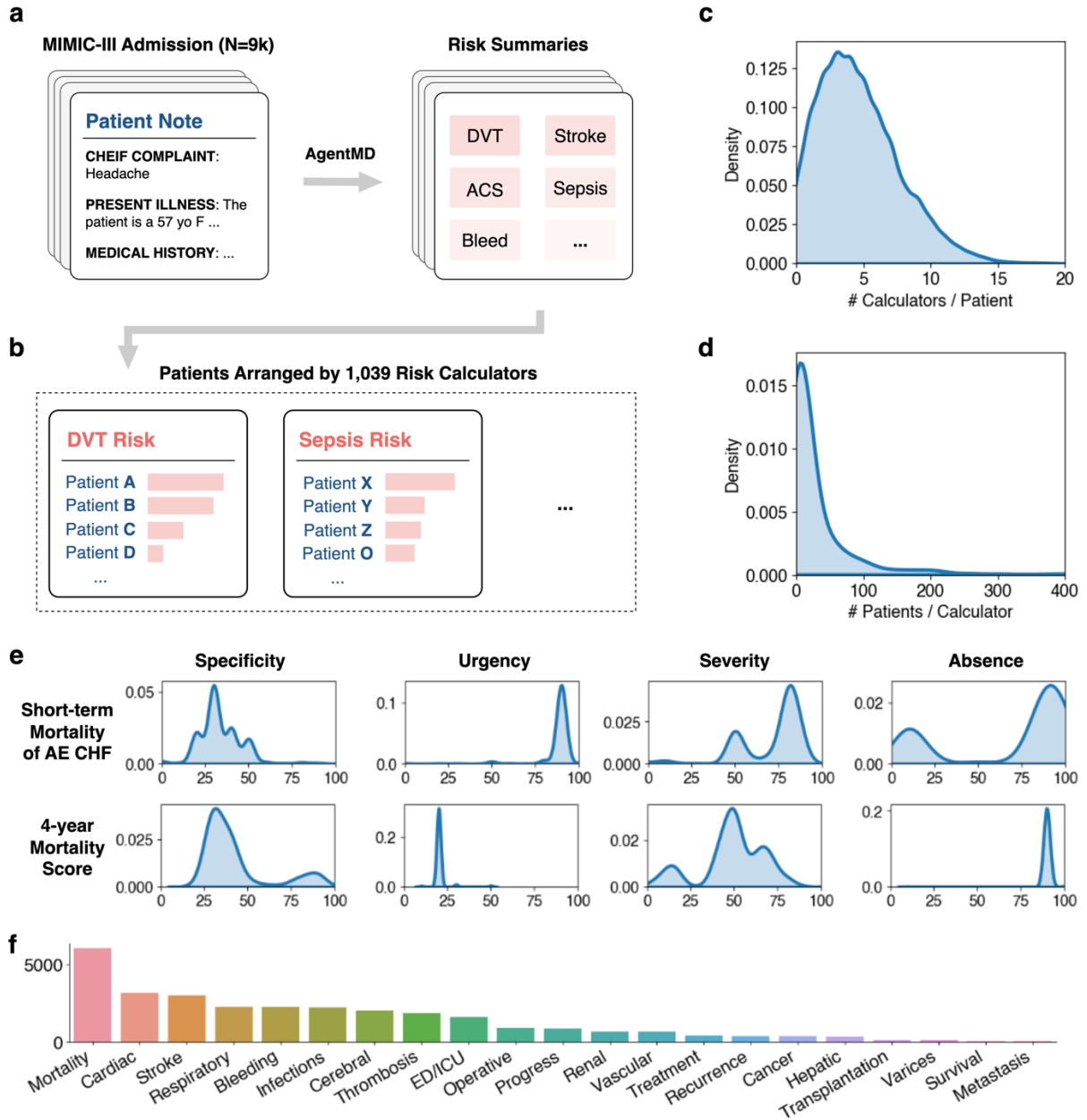

**Fig. 5. Applying AgentMD on the MIMIC-III cohort. a**, AgentMD is applied to 9,822 admission notes in MIMIC. **b**, AgentMD calculation results are aggregated by the risk calculators and patients are ranked within each tool. **c**, The distribution of the number of selected calculators for each patient. **d**, The distribution of the number of eligible patients

for each used calculator. **e**, Calculation result distributions from two calculators. **f**, The number of patients that different types of risk calculators are applied to.

## Discussions

In this study, we aim to address two critical issues in clinical tool learning by the LLMs: the absence of an existing comprehensive clinical toolbox, alongside a deficiency in methodologies and evaluations for tool application. Motivated by the fact that many clinical calculators are shared as free-text in the biomedical literature, such as the CURB-65 score[22], we used PubMed as the knowledge source to curate the clinical toolbox at scale. Exploiting the language and code generation capabilities of the LLMs, AgentMD first makes over 2,000 tools in RiskCalcs by mining the publicly available PubMed abstracts. Our manual evaluations on a representative subset of 100 calculators have demonstrated the high-quality of RiskCalcs, with 67.0% of the calculators evaluated as "Correct" in all three aspects, and 14.0% of the calculators evaluated as "Correct" in two and "Partially Correct" in one aspect. RiskCalcs also covers various clinical tools that have not been implemented elsewhere on the Web, showing its potential as a supplement to the existing clinical calculator hubs such as MDCalcs.

The clinical calculation tools are implemented as reusable Python functions by AgentMD, which results in both generalizability for different downstream LLMs and sufficient precision for computation. More recent efforts such as OpenMedCalc[15] restrict the implementation schema of clinical calculators, which might not be applicable to other

LLMs than GPTs. However, its implementation is instead restricted to the preselected calculators from MDCalc. On the other hand, Almanac[16] implements the use of certain calculators as retrieval-augmented generation with their raw textual descriptions, which are imprecise according to our results in Fig. 4. Our experimental results on RiskQA demonstrate that while GPT-4-based AgentMD achieved the tool selection accuracy of 0.823, the performance decreases to 0.663 with GPT-3.5 as the backbone LLM, indicating room for further improvements. In general, AgentMD with GPT-4 achieved high overall performance on RiskQA (0.877 in accuracy).

Although our research highlights the potential of clinical language agents like AgentMD, it is subject to several limitations. First, the creation of calculator tools was restricted to PubMed abstracts, overlooking detailed descriptions available in full-text articles. Future work should aim to broaden the data sources for tool development. Second, the utilization of GPT-4 as the core LLM in AgentMD introduces a notable limitation due to its high operational costs and the challenge of deploying it locally, which raises concerns regarding data privacy and security. This constraint underscores the potential benefits of investigating alternative, possibly open-source LLMs like Llama[32], which may offer more cost-effective and flexible deployment options while maintaining adherence to stringent data protection standards. Finally, our evaluations were conducted on a relatively small scale, focusing on a sample set of 100 clinical calculators for manual quality assessment and the RiskQA dataset of 350 USMLE-style questions. Future studies should focus on developing more realistic clinical calculation tasks beyond multi-choice accuracy and evaluating the reasoning processes different systems. Larger-scale clinical studies on a diverse set of patients are imperative to validate the effectiveness of AgentMD.

In conclusion, AgentMD represents a promising methical approach to enhancing clinical decision-making through the automated generation and application of a comprehensive set of clinical calculators, RiskCalcs, derived from PubMed articles. Further research is needed to confirm the tool's effectiveness across varied populations and datasets.

## Online Methods

**Backbone LLMs**

In this study, we use GPT-3.5 (model index: gpt-35-turbo) and GPT-4 (model index: gpt-4) from OpenAI as the backbone LLMs for AgentMD and compared baselines, because they have been shown to achieve state-of-the-art performance in both general and biomedical domain[17,33]. We access these models through the Application Programming Interface (API) of Azure OpenAI services, which is compliant with the Health Insurance Portability and Accountability Act (HIPAA). We use the model version 0613 for both GPT-3.5 and GPT-4, and the API version of 2023-07-01-preview. We set the decoding temperature to be 0.0 across this study to ensure deterministic outputs.

**RiskCalcs – AgentMD Tool Creation**

In Supplementary Figure 1, we show a detailed pipeline of building the RiskCalcs tool collection. The first step (Supp. Fig. 1a) is to filter over 37 million PubMed abstracts for potential risk calculators with a Boolean query: *"patient" AND ("risk" OR "mortality") AND ("score" OR "point" OR "rule" OR "calculator")*. About 340 thousand articles passed the initial screen. We then used GPT-3.5 to further screen articles that describe a new risk score or risk calculator with Prompt #1, which leads to about 33 thousand articles for tool creation (Suppl. Fig. 1b).

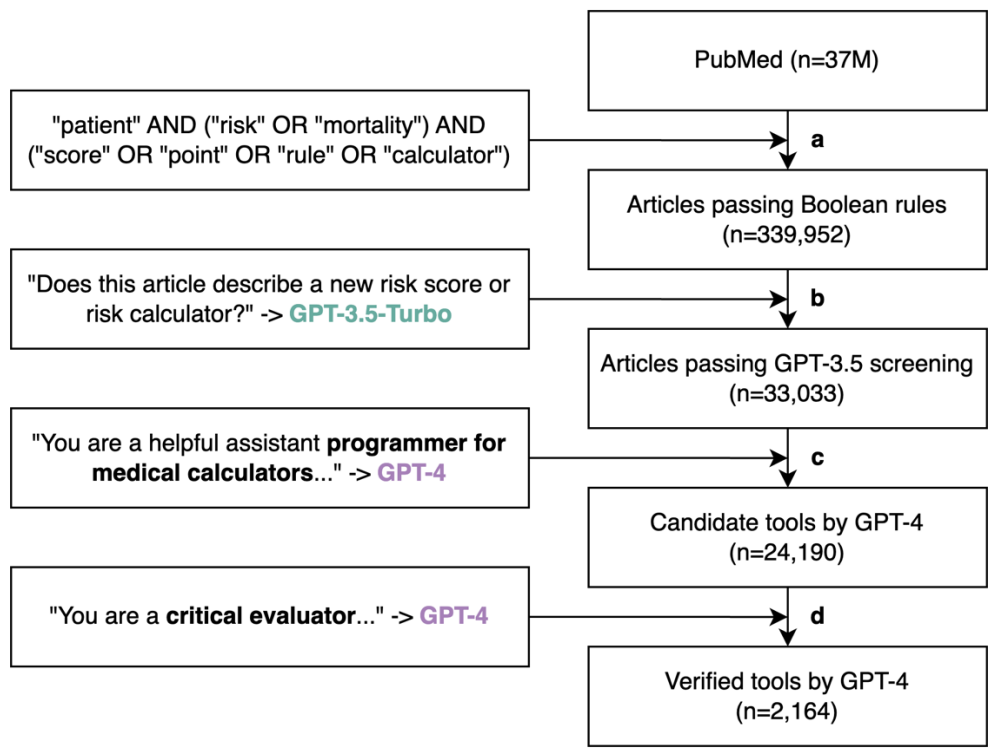

**Supplementary Fig. 1**. The pipeline of building RiskCalcs. **a**, All PubMed abstracts are first filtered by a Boolean query for potential risk calculators. **b**, GPT-3.5-Turbo further screened the articles for ones that describe a new risk score or calculator. **c**, GPT-4 drafts candidate calculators for the articles that have passed GPT-3.5 screening. **d**, The drafted

calculators are then verified by GPT-4, and 2,164 calculators have passed the verification and are included in RiskCalcs.

We used GPT-4 to generate the RiskCalcs tools in two steps: the first step (Supp. Fig. 1c) is to create the candidate tools given the PubMed article with Prompt #2 and one example is shown in Fig. 2a; the second step (Supp. Fig. 1d) is to verify the generated calculators with a list of quality-checking questions (using Prompt #3), and only calculators that passed all verification questions are included in the RiskCalcs collection. Overall, GPT-4 has generated 24,190 tools, where 2,164 tools have passed the verification and have been included in the RiskCalcs collection.

**AgentMD Tool Selection**

As shown in Fig. 2b, there are two-steps in AgentMD tool selection: first-stage retrieval by MedCPT and second-stage selection by LLMs. The goal of first-stage retrieval is to return the top ten most relevant calculators from all potential tools in RiskCalcs. To do this, we first build a Faiss Index[34] of the RiskCalcs collection, where the representation of each calculator is generated using the MedCPT article encoder and the raw PubMed article for the calculator. Each patient is then encoded into a dense vector by the MedCPT query encoder, and the top-10 most similar calculator is returned by searching the RiskCalcs index with exact nearest neighbor search for the dot product metric. In the second stage, the top-10 calculators as well as the patient notes are sent to the LLMs, which then decide which calculators should be used for the given patient.

**AgentMD Tool Computation**

Given a selected tool from RiskCalcs and the patient note, AgentMD uses the calculator by interacting with a Python interpreter, as shown in Fig. 2b. Specifically, AgentMD analyzes the patient information, extracts or infers the needed parameters, and passes them to call the calculator function present in RiskCalcs. The Python Interpreter runs all the code blocks in the AgentMD response and returns either the successful running results or error messages (exceptions in Python). The returned message from the Python Interpreter is then appended to the AgentMD messages, and AgentMD will be prompted again to generate the next analysis or updated code. The tool computation loop will end once AgentMD generates a message that starts with a pre-defined text span "Summary: ", and the paragraph after the summary will be used as the AgentMD Tool Summarization shown in Fig. 2b.

**RiskCalcs Evaluation**

We selected the top-50 most cited calculators and a randomly sampled subset of 50 calculators from the rest of RiskCalcs for manual quality evaluation. The article citation counts were derived from the iCite API (https://icite.od.nih.gov/api) on Dec 17, 2023. Three annotators have been employed to annotate the quality (abstract usefulness, computing logic correctness, and the result interpretation correctness; with the labels of "Correct", "Partially Correct", and "Incorrect") and coverage metrics (implemented by MDCalcs and / or other online implementations returned from the first Google page with the labels of "Covered" or "Not covered"). Initially, the three annotators conducted the annotations independently. Then, they had a round of discussion for the disagreed annotations and determined consensus annotations. One annotator further conducted the computation of five synthetic parameter sets per validated calculator, and compared the computed results with the AgentMD computation results. If the AgentMD computation results are the same as the results from an online implementation (e.g., from an MDCalcs calculator), the label is "Correct (external)". If the results come from the annotator's computing based on PubMed abstracts, the label is "Correct (internal)". Otherwise, the label is "Incorrect".

**RiskQA Evaluation**

RiskQA contains 350 USMLE-style multi-choice question answering instances, where the questions are descriptions of patients and the answer options are different quantitative outcome measures (e.g., mortality). Under the standard RiskQA setting, AgentMD performs tool selection (with MedCPT and Prompt #4), tool computation (with Prompt #5), and finally generates the answer. For the CoT baseline, we appended "Let's think step-by-step" after the whole question as the prompt to LLMs. Under the tool-provided RiskQA* setting, AgentMD directly uses the ground-truth tool for computation, bypassing the tool selection step. The "Name" baseline directly provides LLMs with the name of the ground-truth calculator. Similar to the vanilla RAG setting, the RAG baseline for RiskQA* can directly use the PubMed abstract of the ground-truth tool. Unlike AgentMD, the RAG baseline can only use the raw text description of the calculator and is not connected with a Python interpreter. As such, the performance difference between AgentMD and RAG on RiskQA* reflects the importance of implementing the textual descriptions as functional tools and executing code in medical calculation tasks.

**MIMIC-III Notes**

To preprocess the MIMIC-III dataset, we used the scripts provided by a previous study[34] via https://github.com/bvanaken/clinical-outcome-prediction to get the admission notes. We applied AgentMD to the test split of 9,822 patients. Experiments on the MIMIC-III dataset uses a slightly modified tool selection because various risks should be considered for each patient. As such, AgentMD first uses LLMs to generate a list of potential risk descriptions for the given patient admission note with Prompt #6. Then, AgentMD applies tool selection to each generated risk using Prompt #7, where the patient representation is computed by encoding the textual description of the risks with MedCPT. Finally, AgentMD uses Prompt #8 to conduct the tool computation given the patient note and a selected tool. After AgentMD has computed the results of a given patient-calculator pair, it will further use Prompt #9 to evaluate the specificity, urgency, severity, and absence in the note of the risk results for the patient. These scores are used to draw the calculator result distributions in Fig. 5e.

## Data availability

PubMed abstracts can be downloaded at https://ftp.ncbi.nlm.nih.gov/pubmed/baseline/. The MIMIC-III dataset is available at https://physionet.org/content/mimiciii/1.4/. The

MedCPT encoders are available at https://github.com/ncbi/MedCPT. The Faiss package is available at https://github.com/facebookresearch/faiss. We will publicly release the code and data used in AgentMD upon publication.

## Acknowledgments

This research was supported by the NIH Intramural Research Program, National Library of Medicine. We also thank Charalampos Floudas and Balu Bhasuran for helpful discussions.

## Author contributions

Q.J., Q.C., and Z.L. designed the study. Q.J. conducted the data collection, model construction, model evaluation, and manuscript drafting. Z.W. and Y.Y. carried out the data collection and analysis. Z.W., Y.Y., Q.Z., A.T., D.W., and W.J.W. contributed to the data annotation. Z.L. supervised the study. All authors contributed to writing the manuscript and approved the submitted version.